\documentclass[10pt, a4paper]{article}

\usepackage{lrec}
\usepackage{multibib}
\newcites{languageresource}{Language Resources}
\usepackage{booktabs}
\usepackage{graphicx}
\usepackage{tabularx}
\usepackage{soul}
\usepackage{epstopdf}
\usepackage[utf8]{inputenc}

\usepackage{hyperref}
\usepackage{xstring}
\usepackage{multirow}
\usepackage{todonotes}

\newcommand{\fact}[1]{\begin{footnotesize}{\fontfamily{phv}\selectfont \textbf{#1}}\end{footnotesize}}

\title{Simple Large-scale Relation Extraction from Unstructured Text}

\name{Christos Christodoulopoulos, Arpit Mittal}

\address{Amazon Research Cambridge \\
         \{chrchrs, mitarpit\}@amazon.co.uk\\}
         
\date{}

\begin{document}

\abstract{
Knowledge-based question answering relies on the availability of facts, the majority of which cannot be found in structured sources (e.g. Wikipedia info-boxes, Wikidata). One of the major components of extracting facts from unstructured text is Relation Extraction (RE). In this paper we propose a novel method for creating distant (weak) supervision labels for training a large-scale RE system. We also provide new evidence about the effectiveness of neural network approaches by decoupling the model architecture from the feature design of a state-of-the-art neural network system. Surprisingly, a much simpler classifier trained on similar features performs on par with the highly complex neural network system (at 75x reduction to the training time), suggesting that the features are a bigger contributor to the final performance.\\ \newline 
}

\maketitleabstract

\section{Introduction}
Knowledge-based question answering relies on the availability of facts -- usually in the form of triples, stored in large-scale knowledge bases (KBs) e.g. Freebase \cite{bollacker2008freebase}, DBPedia \cite{auer2007dbpedia}. There are two main sources of facts for such a KB%\footnote{A third source is manually entered knowledge, which is how most large-scale KBs (Freebase, Wikidata, Cyc) are populated.}
: structured data (e.g. Wikipedia info-boxes, Wikidata) or unstructured text. Undeniably, the former type of knowledge extraction is very accurate and has been the main source of knowledge behind the major industrial knowledge bases. However, the facts extracted from structured sources cover a limited set of high-importance relations, leaving a large number of them implicitly (or explicitly) mentioned in unstructured text \cite{mccallum2005information}. 

In order to ground the following presentation, we will present a typical problem from the factual knowledge extraction domain with the following unstructured text from a Wikipedia page:
\begin{quote}
``\textit{Carrie Fisher wrote several semi-autobiographical novels, including Postcards from the Edge.}''
\end{quote}
The purpose of a fact extraction system is to extract the following facts of the form of \texttt{predicate} (\textit{subject}, \textit{object}): \fact{instance of (postcards from the edge, novel)}, and \fact{author of (postcards from the edge, carrie fisher)}, where the first part is a \textit{relation}, and the other parts are the left and right \textit{entities} participating in that relation. 

Typically three tasks are involved in generating facts: Entity Recognition, Entity Resolution (or Entity Linking), and Relation Extraction (RE). Entity Recognition and Resolution deal with the task of translating surface strings to KB entities. This includes nominal or pronominal coreference resolution: we should be able to extract the same entity even if the text stated that `Fisher wrote\ldots' (instead of resolving e.g. to Bobby Fisher) or `She wrote\ldots' (provided that Carrie Fisher's name was mentioned in a previous sentence). Relation Extraction extracts relation triples (or facts) involving those entities with appropriate relations (also part of the KB schema). Each of these components could be built and operated in isolation, but they affect the performance of each other.

In this paper, we examine the task of RE focusing on extracting knowledge to enrich a large-scale KB ($\sim$billions of facts). We consider a state-of-the-art model that has been applied to hyponymy detection and present a thorough analysis of its application to datasets derived from Wikidata and Alexa KB, a proprietary large-scale triple KB that powers Amazon's Alexa. We also present a new way of generating distant supervision for relation extraction with a simple yet effective way of reducing the noise for the entity resolution.

\section{Related work}\label{sec:relatedwork}
Relation Extraction is the NLP task of extracting structured semantic relations between entities from natural (unstructured) text. Formally, it can be defined as identifying semantic relations between (resolved) entities and normalise these relations by mapping them to a predefined KB schema. In the NLP community, the RE task evolved out of the Information Extraction projects like MUC in the 1990s (see \newcite{chinchor1993evaluating} for an overview) and ACE in the 2000s \cite{doddington2004automatic}. In both projects the main focus was the automatic extraction of \textit{events} rather than relations (the main difference being that an event is a special type of fact that involves actor entities and occurs at a specific time point) %\footnote{The main difference between an event and a fact is that an event is a special type of fact that involves actor entities and occurs at a specific time point.} 
in a limited set of domains (e.g. bombings, company mergers, etc.). This meant that in both projects the number of relations marked for extraction was very limited (3 relations in MUC and 24 in ACE with 7k relation instances for 40k entity mentions).

Starting with those projects, the task of RE was thought of as a pipeline, where the entities were first detected, resolved to a standard schema, and then the RE system would determine which of the possible relations was expressed (if any) between any given pair of entities. Much of the earlier work explored a variety of different features, such as syntactic phrase chunking and constituency parsing \cite{bunescu2005shortest,jiang2007systematic,qian2008exploiting}, and semantic knowledge like WordNet \cite{zhou2005exploring}, although \newcite{jiang2007systematic} showed that the more complex features might actually hurt the performance of an SVM-based RE system. The work of \newcite{shwartz2016improving}, that we closely follow, is also using both semantic and syntactic features, by combining the dependency paths between entities, with word embedding representations of both the entities and the lemmas in the dependency paths.

% Another related area is joint entity detection/resolution and relation extraction \cite{roth2007global,singh2013joint,li2014incremental,pawar2017end}. We also need to perform an entity resolution step, but we are using a deterministic gazetteer lookup solution, rather than a learned model.
Another related area is relation extraction for Open Information Extraction (OpenIE). Some of the more representative projects in the area, like Reverb \cite{fader2011identifying} and more recently ClauseIE \cite{del2013clausie} use syntactic information (PoS tagging / chunking, and dependency parsing respectively) to extract entity and relation phrases. However, unlike OpenIE, we are interested in normalized entities and relations (i.e. that map to a knowledge base).

In this work, we follow a common way of producing training examples for RE is to use \textit{distant supervision} \cite{craven1999constructing,mintz2009distant}: the assumption is that if any sentence mentions two entities which we know (from a KB) participate in a specific relation, that sentence must be evidence for that relation. In the area of distant supervision, there are two relevant research directions. The first is to use it for directly enriching KBs from unstructured text, as well as leverage the KBs to generate the distant supervision labels \cite{poon2015distant,parikh2015grounded}. The second direction attempts to reduce the noise in distant supervision labels. A first line of approaches, starting with Data Programming \cite{ratner2016data}, uses generative models to combine multiple sources of weak supervision (e.g. automatically extracted from a KB, rules generated by experts etc.) in order to predict disagreements and overlaps between them and create a noise-aware posterior distribution of predictions. An extension of this approach is Socratic Learning \cite{varma2017socratic} which uses the differences in the predictions of the generative model and the main classification system to discover discriminating features and add them back to the generative model. As these approaches require multiple sources of weak supervision, we examine another line of projects which works by aggregating the support sentences\footnote{By support sentences we mean any sentence in the dataset that contains both entities.} for each entity pair \cite{riedel2010modeling,hoffmann2011knowledge}. This is the approach that \newcite{shwartz2016improving} and the current work follow.

\begin{figure*}
\centering
	\includegraphics[width=.9\textwidth]{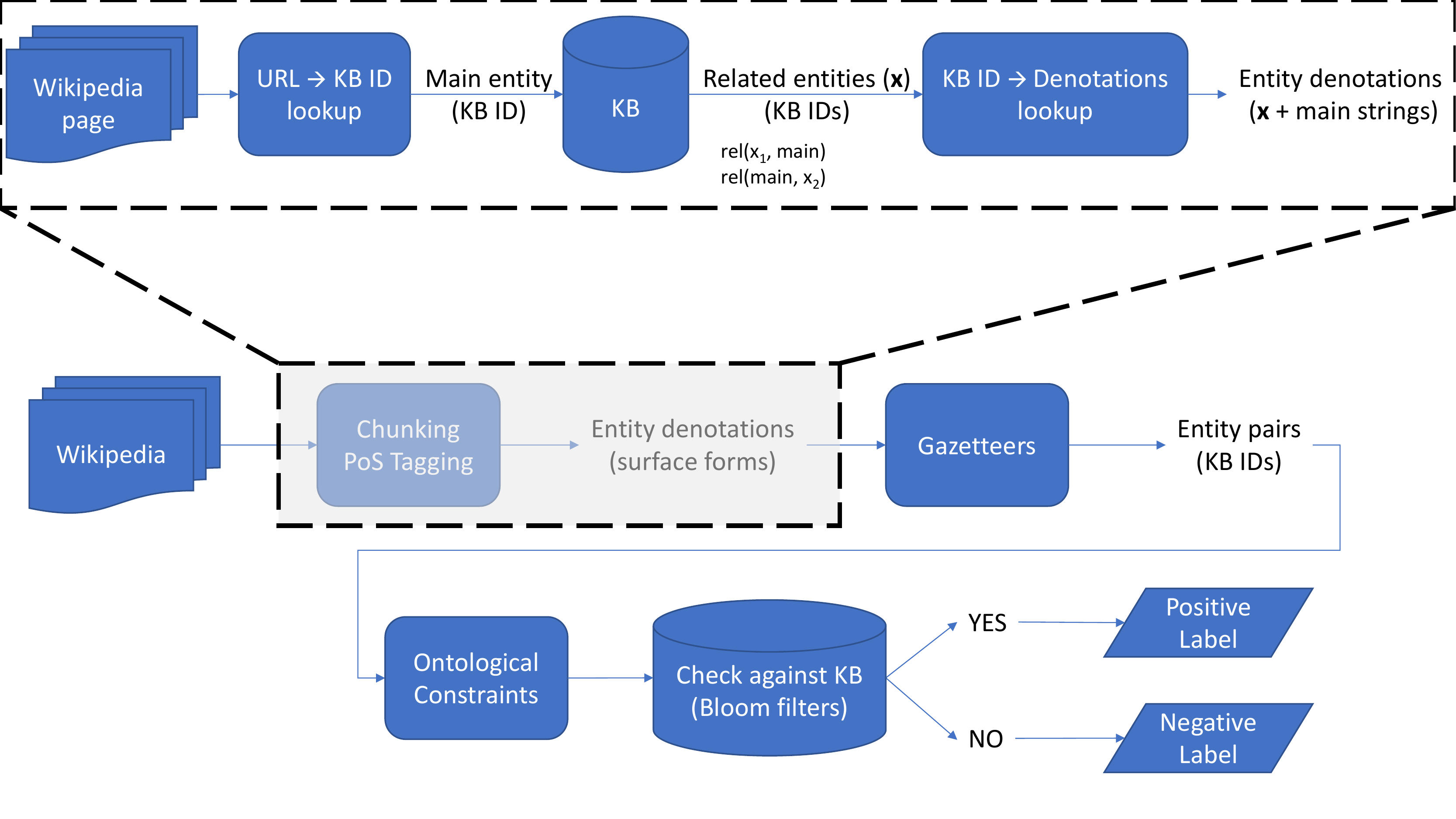}
	\caption{The distant supervision pipeline with page-specific gazetteers. The grey box represents the entity resolution system of \protect\newcite{shwartz2016improving}.}
	\label{fig:dist-super-pipeline}
\end{figure*}

\begin{figure*}
\centering
\includegraphics[width=.9\textwidth]{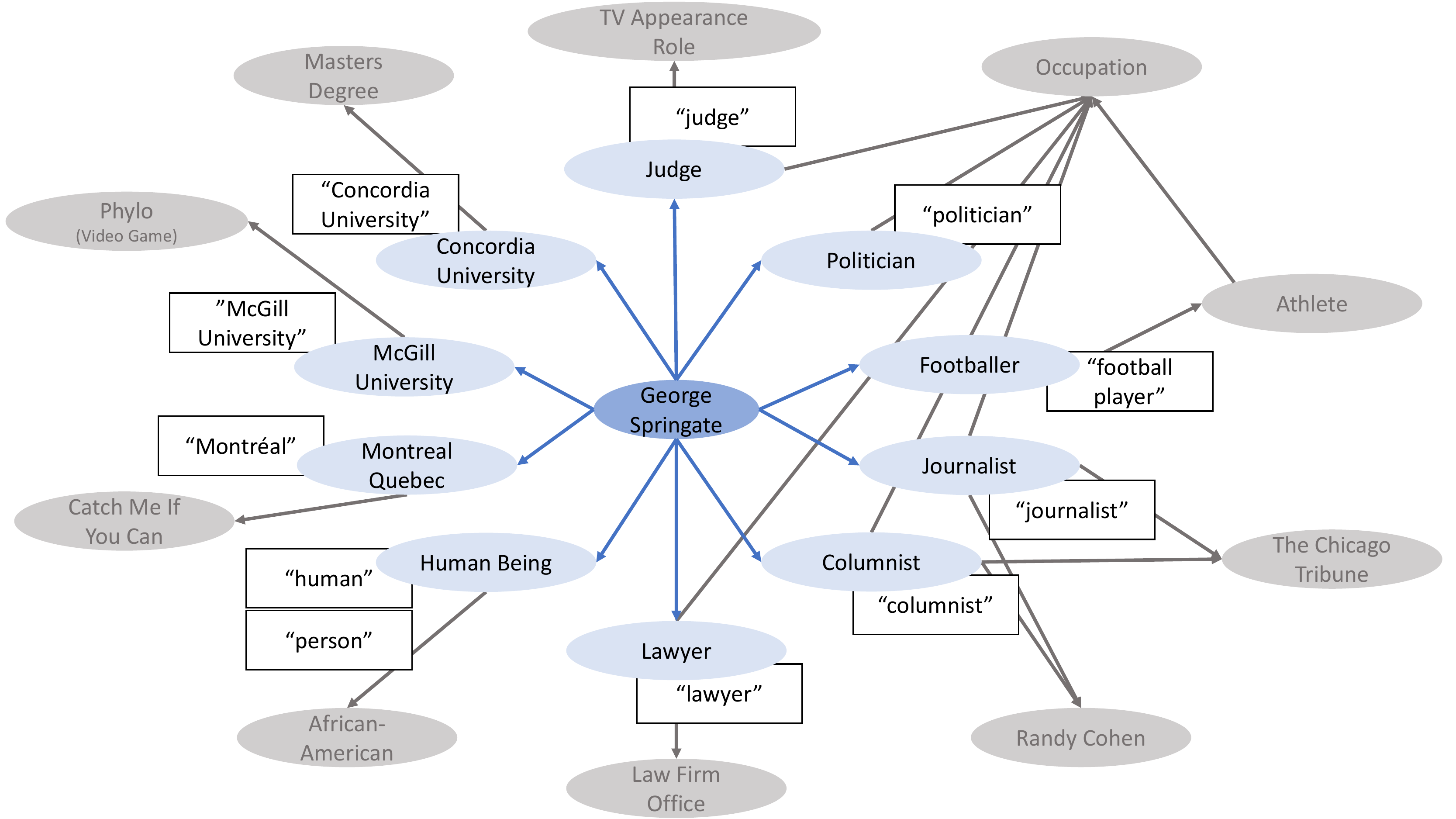}
\caption{Example KB entry for George Springate. The lightly (blue) shaded ovals represent entities that are within one hop from the main entity and the white boxes are their denotations. The grey ovals represent entities that are two hops away.}
\label{fig:springate}
\end{figure*}

\subsection{HypeNET} \label{sct:hypenet}
A recent paper \cite{shwartz2016improving} proposed HypeNET, a new method for RE that integrated dependency path information with distributional semantic vector representation of the entities. The authors applied this method to extract hyponyms (i.e. \fact{instance of} relations) and also made a new version of their system publicly available.\footnote{\url{https://github.com/vered1986/LexNET}} The training examples used (entity/relation triples) come from a number of sources like WordNet \cite{miller1995wordnet}, Yago \cite{hoffart2013yago2}, DBPedia and Wikidata, and the source of the linguistic features (part-of-speech tags, dependency paths, noun phrases) was the 2015 dump of Wikipedia, processed using the spaCy system\footnote{\url{https://spacy.io}}.
Their proposed system achieved by far the best results on their dataset. Since \fact{instance of} is one of the most often used relations (most of the uses are implicit, during inference), we decided to investigate HypeNET as the base of our RE system. 

The training examples used by the authors of HypeNET consisted of facts about only one relation. We wanted to build a system that works on multiple relations at a very large scale. Hence, in this work we use two different dataset sources: Wikidata, a publicly-available large-scale KB to aid reproducibility, as well as the larger Alexa KB, built by combining a hand-curated ontology with publicly available data from Wikidata, Wikipedia, Freebase, DBPedia, and other sources.
%Hence, in this work we used a proprietary KB (citation will be added in the final version), built by combining a hand-curated ontology with publicly available data from Wikipedia, Freebase, DBPedia, and other sources. For comparison with other KBs, we have roughly 20k classes, 2k properties or relations, and on the order of 1 billion facts for 50 million entities. Compared to the 70k entity pairs in the HypeNET data, our KB contains nearly 52 million facts for the \fact{instance of} relation (however, for many of these facts there might not be a single Wikipedia sentence that supports them).

\section{Distant supervision} \label{sct:dist-supervision}
Following the technique presented in \newcite{mintz2009distant}, and the implementation in HypeNET, we needed to generate training examples where entities $X$ and $Y$ are connected by a relation in the KB and also appear together in the same sentence. When we applied the distant supervision technique presented in HypeNET to our datasets (both Wikidata and Alexa KB) we got poor annotations (see Figure~\ref{fig:er-examples}(top) for some examples from Alexa KB and section \ref{sct:results-dist-supervision} for evaluation on both datasets). This could be attributed to the large volume of entities and their corresponding denotations in the KBs, which resulted in a number of ambiguous situations. For instance, ``\textit{Chicago}'' could denote both the \textbf{city} and the \textbf{broadway musical show}. In the following section we present our new technique of filtering denotations used for Entity Resolution. This method allows our RE system to scale much better than the original method.

\subsection{Page-specific gazetteers} \label{sct:page-specific-gazetteer}
We created a new type of entity gazetteer, based on the main entity of a Wikipedia page, and the knowledge about that entity we have in the KB. The new system, presented on the top dashed box in Figure \ref{fig:dist-super-pipeline}, starts with a Wikipedia URL, retrieves its corresponding ID from the KB for that URL (the main entity), and then extracts entities that are connected directly to the main entity (one-hop distance in the KB graph), by going through all the relations the main entity is involved in (except those involving string literals) and returning the entities on the other side of those relations. For each of the related entities, we collect its denotational strings into a purpose-built gazetteer. Figure \ref{fig:springate} shows an example KB subgraph for a target entity (in this case George Springate); it contains all the entities immediately connected to it with relations such as \fact{graduate of} or \fact{instance of}. Also appearing in the graph, are the denotation strings for each one of the related entities.

Note that this approach will reduce the number of extracted entities compared to the original method, but will dramatically improve both the coverage for non-NP entities and precision of entity resolution. One way to increase the recall of this system would be to consider entities with a distance of $>$1 (entities related to entities related to the main entity). Figure~\ref{fig:er-examples} (bottom) shows results obtained by performing entity resolution using page-specific gazetteers. Those examples, as well as the results in section \ref{sct:results-dist-supervision} show that the noise in the data is significantly reduced.

\subsection{Annotation pipeline}
The bottom half of Figure \ref{fig:dist-super-pipeline} presents the distant supervision generation process adapted from \newcite{shwartz2016improving} to work with our data. In the original work, the text is processed to split and tokenise the sentences, tag the parts of speech and separate the noun phrases (NPs) -- these are the candidate entities. They then construct the dependency path between each possible pair of entities. Each noun/NP pair is checked against the KB for distant supervision.  keep only the entities/paths that appear in the list of labelled examples. They also filter out entity pairs that have infrequent paths (occurring fewer than five times), and pairs whose path is more than five tokens long. However, as discussed in the beginning of this section, this approach introduces a lot of noise. To avoid this problem, we use the page-specific gazetteer and a greedy string matching system to scan through the unstructured text and assign KB IDs to the longest-matching substring in a sentence.

The final step was to generate the annotation labels themselves. To do that, we examine each possible pair of entities to see if they participate in the target relation. For the Wikidata KB, we simply checked whether the target relation existed as a property in the data. Considering the large size of Alexa KB, database lookup operations could be very expensive. In order to speed up the lookup for each $X~rel~Y$ triple, we used two methods. First, before checking against the KB though, we ensure that the pair conforms with the class signature of the relation (`Ontological Constraints'). For example, only a \fact{geographical location} can be the left entity in the \fact{birthplace of} relation. Second, instead of relying on database queries, we used Bloom filters \cite{bloom1970space} -- a memory efficient probabilistic data structure that can be used to test if an entity is a member of a set. The compression value of a Bloom filter is governed by the accepted false positive rate. We set the false positive rate to 0.001 for our experiments. %Bloom filters are used instead the `Check against KB' stage in the Wikidata case. 

Since for any given pair of entities it is much more likely that they are not going to be related, we only keep a small fraction of the negative instances. Following 
%\newcite{mintz2009distant} and 
\newcite{shwartz2016improving}, we use a 4:1 negative to positive ratio.

\section{Isolating HypeNET features}\label{sct:isolating}
To discover the effectiveness of the approach of~\newcite{shwartz2016improving}, we wanted to separate HypeNET's neural architecture from its input features and use those features with different (and simpler) classifiers. HypeNET's main advantage is that it integrated dependency path features with distributional information about the word lemmas along the path and left and right entities. As our goal was to generate discrete features to be used with more traditional classifiers, we opted for using Brown clusters \cite{brown1992class} instead of the 50-dimensional GloVe vectors \cite{pennington2014glove} used by \newcite{shwartz2016improving}. The Brown clusters were pre-trained on the Reuters Corpus Vol. 1 \cite{lewis2004rcv1} using 3,200 clusters.

After evaluating different feature configurations (see section \ref{sct:results-features}), the resulting features were as follows: for each entity pair and for each support, we extracted the dependency path between them and concatenated the lemma, 4-bit prefix of Brown cluster of the lemma, part of speech, dependency relation, and path direction information; to that we added the strings and 4-bit Brown cluster prefix of the left and right entities. The features from different supports were concatenated into one feature list. For example, given the following sentences containing the entity pair \fact{carrie fisher, star wars}: ``\textit{In 1977, Fisher starred in George Lucas' film Star Wars}'', and ``\textit{Fisher became known for playing Princess Leia in the Star Wars film series}''. The following is the full list of discrete features extracted, where each space-separated token is a distinct feature, and X and Y are used to replace the left and right entities:\\
\begin{footnotesize}Carrie\_Fisher/0111 X/0000/NOUN/nsubj/$>$ star/0011/VERB/ROOT in/1101/ADP/prep/$<$ film/0010/NOUN/pobj/$<$ Y/0000/NOUN/appos/$<$ X/0000/NOUN/nsubj/$>$ become/1111/VERB/ROOT know/1111/VERB/acomp/$<$ for/1101/APD/prep/$<$ play/1111/VERB/pcomp/$<$ in/1101/APD/prep/$<$ Y/0000/NOUN/pobj/$<$ \hfill Star\_Wars/0011\end{footnotesize}

\begin{figure}
\centering
	\includegraphics[width=.9\columnwidth]{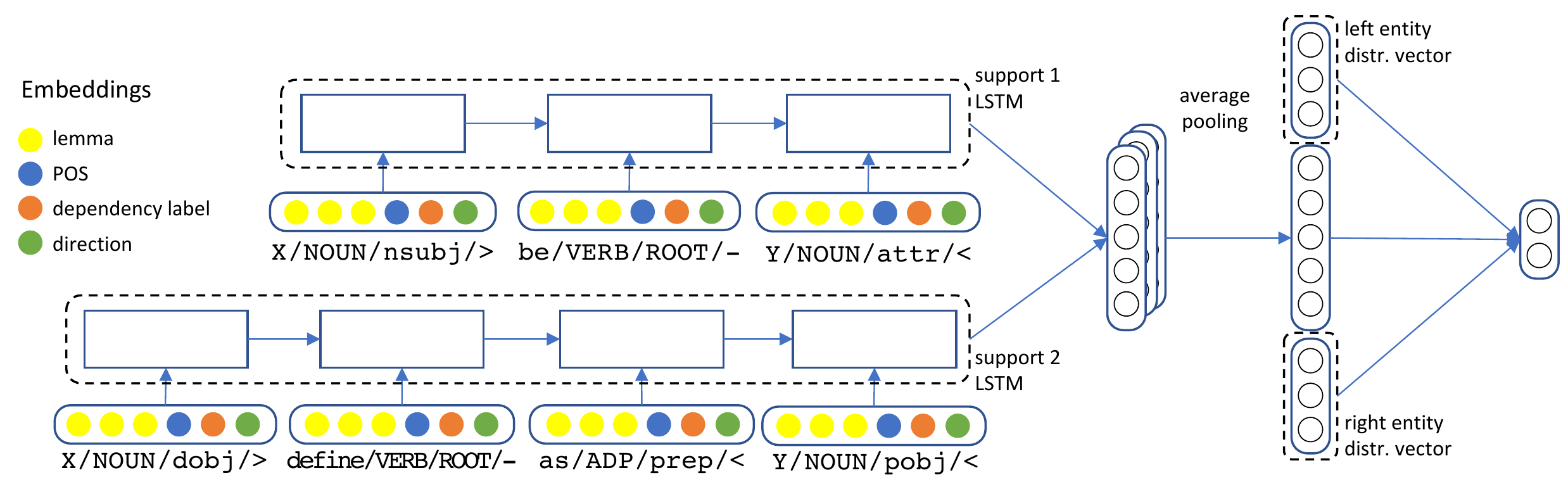}
	\caption{The HypeNET model architecture, reproduced from \protect\newcite{shwartz2016improving}.}
	\label{fig:hypenet-arch}
\end{figure}

\subsection{Using a MaxEnt classifier}
In the first set of experiments, we used a standard Maximum Entropy classifier from MALLET toolkit \cite{mccallum2002mallet} with the discrete features described above. The parameters and settings were kept to their defaults (LBFGS optimizer, with a Gaussian prior variance of 1).

\begin{figure*}
\centering
\framebox{\parbox{15.5cm}{
His studies were interrupted by army service and at the \textit{end} of the \textit{war} he was forced to return\ldots\\
\fact{instance of (the second world war, cause of death)}

\vspace{0.7em}
In the \textit{intro} to the \textit{song}, Fred Durst makes reference to\ldots\\
\fact{instance of (intro 15367, song)}

\vspace{0.7em}
Turner also released one \textit{album} and several \textit{singles} under the moniker Repeat.\\
\fact{instance of (the singles the 2011 album, album)}

\vspace{0.5em}
\noindent\rule{14cm}{0.2pt}

\vspace{0.5em}
\textit{Call Your Girlfriend} was written by Robyn, Alexander Kronlund and Klas Åhlund, with the latter producing the \textit{song}.\\
\fact{instance of (call your girlfriend 3, song)}

\vspace{0.7em}
\textit{Forget Her} is a \textit{song} by Jeff Buckley.\\
\fact{instance of (forget her, song)}

\vspace{0.7em}
The \textit{Subei Mongol Autonomous County} is an autonomous \textit{county} within the prefecture-level city of Jiuquan in the northwestern Chinese province of Gansu.\\
\fact{instance of (subei mongol autonomous county, chinese county)}
}}
\caption{Entity resolution results for the distant supervision training data using Alexa KB and the original pre-processing system of \protect\newcite{shwartz2016improving} (top), and the new page-specific gazetteers (bottom). The matched strings in the original sentences are \textit{highlighted}.}
\label{fig:er-examples}
\end{figure*}

\subsection{Using the fastText model}
\newcite{joulin2016bag} recently introduced fastText: a very efficient classifier composed of a simple linear model with a rank constraint. The architecture of the system is very similar to that of \newcite{mikolov2013efficient} except that instead of predicting the middle word in a window, the classifier is predicting a label. For fastText, the input features are token ngrams which are embedded into a single hidden value and fed into a hierarchical softmax classifier. For our experiments, we used fastText's default settings, except for the number of ngrams, which we set to 4.

\subsection{Using the HypeNET model}
The original version of HypeNET (Figure \ref{fig:hypenet-arch}) combines the dependency path-based features with the distributional information in its neural net architecture: for each entity pair, each support (dependency path) token is encoded by a set of embedding layers -- one for each linguistic component -- and passed into an LSTM layer. The LSTM layers for the whole path are merged by an average pooling layer and the distributional representation of the entities (via embedding layers) is added. Finally, a softmax layer makes a binary classification decision.

We implemented our own version of HypeNET code using Keras~\cite{chollet2015keras} and optimized the learning objective using the Adam optimizer. We modified the basic HypeNET model by making the following changes: i) we allowed the training of word embeddings for lemmas (after initializing them with GloVe embeddings), ii) we replaced the uni-directional LSTM with bi-directional LSTM\footnote{The performance of the resulting model was slightly better on the Alexa KB dataset, achieving an F-score value of 94.29$\pm 0.21$, compared to 94$\pm 0.15$ for the basic model across the three trials (with a threshold value of 0.5) for \fact{instance of} relation.}.

\begin{table*}
\centering
\begin{tabular}{@{}rl|cc|cc|c@{}}
\toprule
~ & ~ & \multicolumn{2}{c|}{\textbf{HypeNET}} & \multicolumn{2}{c|}{\textbf{fastText}} & \textbf{MaxEnt} \\
~ & Relation & $\mu$(F-score) & $\sigma$ & $\mu$(F-score) & $\sigma$ & F-score \\
\midrule
\parbox[t]{2mm}{\multirow{3}{*}{\rotatebox[origin=c]{90}{Wikidata}}} & \fact{instance of} & 93.90 & 0.21 & \textbf{96.44} & 0.01 & 58.45 \\
~ & \fact{birthplace of} & 92.06 & 0.90 & \textbf{93.05} & 0.07 & 66.72 \\
~ & \fact{part of} & 48.73 & 2.59 & \textbf{72.87} & 0.16 & 45.13 \\
\midrule
\parbox[t]{2mm}{\multirow{3}{*}{\rotatebox[origin=c]{90}{Alexa}}} & \fact{instance of} & 94.29 & 0.21 & 94.31 & 0.03 & 83.93 \\
~ & \fact{birthplace of} & 85.57 & 0.26 & \textbf{87.63} & 0.01 & 80.83 \\
~ & \fact{applies to} & 81.98 & 1.78 & \textbf{86.17} & 0.01 & 65.27 \\
\bottomrule
\end{tabular}
\caption{Mean and standard deviation of the F-score values at 0.5 threshold across three trials for the Wikidata and Alexa KBs, using the MaxEnt, fastText, and HypeNET (our Keras implementation with word embeddings training and bi-directional LSTMs). The MaxEnt model did not have any variance across the trials.}
\label{tbl:results-all-rels}
\end{table*}

\section{Evaluation}
We want to examine a varied set of connections between the left and right entities, so in addition to the \fact{instance of} relation (P31 in Wikidata) that connects objects to classes, we will examine \fact{birthplace of} (P19) that connects a location entity to a person entity, and \fact{part of} (P527) which links objects to their meronyms. When evaluating against Alexa KB, we replaced \fact{part of} with \fact{applies to}, a relation that links an attribute to an object and has no correspondence in Wikidata. We will use the Wikidata KB as a first source of evaluation, and switch to the Alexa KB for a more in depth exploration.

We evaluate all models on a sample of 50K examples for training, 10K examples for validation and test respectively for all relations (except \fact{part of} for which we could only collect 22K training examples). Each example is the collection of all the sentences supporting a $X~rel~Y$ triple that have been annotated by the distant supervision system of section \ref{sct:dist-supervision}. We examine the effect of grouping supports in section \ref{sct:results-grouping}.

\section{Results and discussion}
% We implemented our own version of HypeNET code using Keras~\cite{chollet2015keras} and optimized the learning objective using the Adam optimizer. We modified the basic HypeNET model by making the following changes: i) we allowed the training of word embeddings for lemmas (after initializing them with GloVe embeddings), ii) we replaced the uni-directional LSTM with bi-directional LSTM\footnote{The performance of the resulting model was slightly better on the Alexa KB dataset, achieving an F-score value of 94.29$\pm 0.21$, compared to 94$\pm 0.15$ for the basic model across the three trials (with a threshold value of 0.5) for \fact{instance of} relation.}.

We ran each of the following experiments three times (with random initialization) to obtain a measure of variance for their results.
 
\subsection{Distant supervision}\label{sct:results-dist-supervision}
The goal of the method presented in section \ref{sct:page-specific-gazetteer} was to reduce the number of false positives at the cost of introducing some amount of false negatives (due to missing entities, missing denotations, or missing KB facts). In order to quantify the effect of the new method, we manually annotated 1,000 \fact{instance of} distant supervision examples produced by our new method and the original method used by \newcite{shwartz2016improving}. The original method yielded 67\% false positive and 3\% false negative examples; the page-specific gazetteer solution returned only 1\% false positives and 39\% false negatives. After more analysis, 62\% of the false negatives (or 24\% of the total examples) were cases were the KB contained the \fact{subclass of} relation, which we consider a separate relation (although in the data collected by \newcite{shwartz2016improving} from Yago and Wikidata it is conflated with \fact{instance of}). The results are similar when using the Wikidata KB: around 1\% false positives but only 5\% false negatives\footnote{The lower rate of false negatives can be partially attributed to the exact lookup, instead of the Bloom filters used with Alexa KB.} of which 89\% were cases where the KB contained similar relations (like \fact{occupation} for people, or \fact{taxon} for species).

Figure \ref{fig:er-examples} presents a qualitative comparison of the two methods on our KB. We can see that the two problems of spurious entity matching (e.g. ``\textit{end}'' to \fact{cause of death}) and non-standard noun-phrase entities (e.g. ``\textit{call your girlfriend}'') have been successfully addressed by the page-specific gazetteer.

\subsection{Model comparison}
The results comparing the performance and generalizability of the models (over the three relations) are shown in Table \ref{tbl:results-all-rels}. The main takeaway from these results is that the more advanced architecture of HypeNET does not offer a significant advantage over that of fastText when used with (almost) the same input features. As an added benefit, the fastText classifier is dramatically faster than the HypeNET model, with a reduction of training time from around 75 minutes to less than a minute. However as the results of the MaxEnt model show, the features alone are not enough. It is fastText's (and HypeNET's) ability to create higher-dimensional representations of these discrete features that provide the best results.

% Figure \ref{fig:HypeNet-examples} presents a sample of the results for the three different relations.

% \begin{figure*}
% \framebox{\parbox{15.5cm}{
% On 22 November 2008, \textit{Costa Concordia} suffered damage to her bow when high winds over the Sicilian city of Palermo pushed the \textit{ship} against its dock.\\
% \fact{instance of (costa concordia, ship)}

% \vspace{0.7em}
% On October 12, 2006, the United States Postal Service building in \textit{Green River}, Wyoming, was officially designated as the Curt Gowdy Post Office Building, honoring the place of \textit{Gowdy}'s birth.\\
% \fact{birthplace of (green river, curt gowdy)}

% \vspace{0.7em}
% \textit{Debora Seilhamer} (born October 4, 1985) is a \textit{female} volleyball player from Ponce, Puerto Rico.\\
% \fact{applies to (female, debora seilhamer)}
% }}
% \caption{Relation extraction results using the fastText system trained with the page-specific distant supervision data.}
% \label{fig:HypeNet-examples}
% \end{figure*}

\subsection{Training data size}
Another parameter we wanted to explore was the impact of size of the training data since we plan to target relations with fewer training examples in the future. We evaluate the F-scores of the HypeNET, fastText, and MaxEnt models for the \fact{instance of} relation on the Alexa KB dataset.
\begin{figure}
\includegraphics[width=\columnwidth]{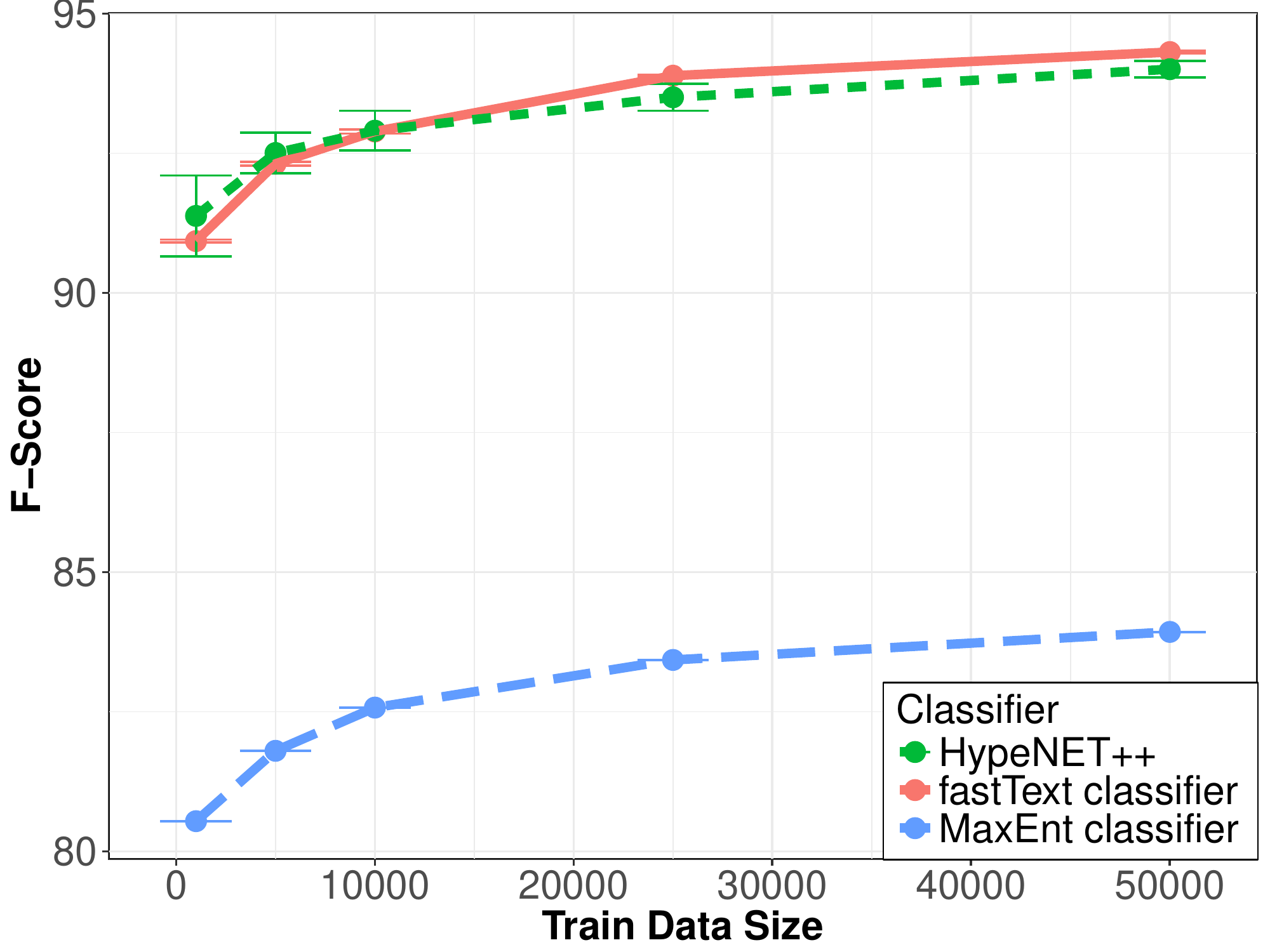}
\caption{Effect of training data size on the Alexa KB dataset for the \protect\fact{instance of} relation for HypeNET and fastText models at 0.5 confidence threshold. %The results of the MaxEnt model are not shown here since they are much lower but follow a similar trend.
}
\label{fig:results-train-size}
\end{figure}

The results are shown in Figure \ref{fig:results-train-size}. Note that the numbers in the figure refer to entity pairs, not individual supports (sentences). As expected, the performance of all systems keeps increasing when more training examples are used, but there are two interesting observations to be made. The first is the relative variance of the fastText versus the HypeNET model, especially for the case of fewer than 25k examples. The second is that even with 1,000 training examples, the F-score of both HypeNET and fastText models is above 90\%.

\subsection{Grouping supports}\label{sct:results-grouping}
We also wanted to investigate the effect of grouping the supports (sentences) for each entity pair. As mentioned earlier (Section~\ref{sec:relatedwork}), this had been proposed as a method to reduce noise in the distant supervision labels \cite{hoffmann2011knowledge}. In \newcite{shwartz2016improving}, the grouping was performed by the mean pooling layer; in the case of the fastText-based system, we simply concatenate the feature tokens from all the supports and feed them into the single hidden layer.

For each of the three relations, we ran the fastText model with and without grouping each entity pair's supports, using exactly the same features in both cases. The Table \ref{tbl:results-grouping} presents the results. Interestingly, the effect on \fact{instance of} is much smaller than on the other two relations. One possible explanation could be that the page-specific gazetteer method is producing fewer false positives for that relation; more likely, the supports for \fact{birthplace of} and \fact{applies to} are more diverse than those of \fact{instance of}, making their grouping more useful to the classifier. 

\subsection{Using satellite nodes}
We also looked at the role of the dependency path \textit{satellite nodes} (words to the left and right of the entities). This type of features has also been adopted by various systems including \newcite{mintz2009distant} and \newcite{shwartz2016improving}, and we wanted to establish a basis for its effectiveness across multiple relations. The results, shown in Table \ref{tbl:results-satellites}, show that the clearest advantage of the satellite nodes is for the \fact{birthplace of} relation; for the other two the performance without satellite notes is marginally different. This suggests that the immediate context of the left and right entities is less informative for the \fact{instance of} and \fact{applies to} relations, possible because of the more limited ways that the expression of these relations are syntactically constructed.

\begin{table}
\centering
\begin{tabular}{@{}lrrr}
\toprule
 & \fact{inst. of} & \fact{appl. to} & \fact{bp of} \\
\midrule
grouped & 94.31 & 86.17 & 87.63\\
ungrouped & 93.85 & 80.90 & 85.09 \\
\bottomrule
\end{tabular}
\caption{Effect of grouping supports for each $X~rel~Y$ triple using the fastText classifier on the Alexa KB data (threshold of 0.5).
}
\label{tbl:results-grouping}
\end{table}

\begin{table}
\centering
\begin{tabular}{@{}lrrr}
\toprule
 & \fact{inst. of} & \fact{appl. to} & \fact{bp of} \\
\midrule
satellites & 94.31 & 86.17 & 87.63 \\
w/o satellites & 93.69 & 85.85 & 84.42 \\
\bottomrule
\end{tabular}
\caption{Effect of using dependency path satellite nodes for each $X~rel~Y$ triple using the fastText classifier on the Alexa KB data (threshold of 0.5).
}
\label{tbl:results-satellites}
\end{table}

\begin{table}
\centering
\begin{tabular}{@{}lrrr}
\toprule
 & \fact{inst. of} & \fact{appl. to} & \fact{bp of} \\
\midrule
(1) 5 supports & 94.55 & 86.26 & 87.56 \\
all supports & 94.33 & 85.92 & 87.63 \\
\bottomrule
\end{tabular}
\caption{Effect of using all the supports for each $X~rel~Y$ triple using the fastText classifier on the Alexa KB data (threshold of 0.5).}
\label{tbl:results-supports}
\end{table}

\begin{table}
\centering
\begin{tabular}{@{}lrrr}
\toprule
 & \fact{inst. of} & \fact{appl. to} & \fact{bp of} \\
\midrule
(1)-Brown & 94.20 & 85.93 & 87.51 \\
(1)-lemma & 94.17 & 84.15 & 86.65 \\
(1)-POS & 94.15 & 85.93 & 87.71 \\
(1)-dep & 93.59 & 85.42 & 86.53 \\
(1)-$X$/$Y$ entities & 93.63 & 83.89 & 86.95 \\
\midrule
$X$/$Y$ only & 91.15 & 74.20 & 81.15 \\
full sentence & 86.70 & 77.77 & 87.09 \\
\bottomrule
\end{tabular}
\caption{F-score results for the three relations on the Alexa KB dataset. The baseline system (1) is the fastText classifier using the 5 most frequent supports for each $X~rel~Y$ triple, (1)-dep refers to the system with both the dependency relation and direction features removed, the last system uses all the (lowercased) words in each support as features.}
\label{tbl:results-ablation}
\end{table}

\subsection{Using all supports} \label{sct:results-all-supports}
A comparison is made for the fastText models trained using the 5 most frequent supports for each triple with the ones trained using all available supports. As shown in Table \ref{tbl:results-supports}, reducing the supports to the most frequent ones slightly increases the performance (except in the case of \fact{birthplace of}) even though on average more than 18K training examples, and more than 3K test examples contain more than 5 supports.

\subsection{Feature ablation}\label{sct:results-features}
As a final step in our exploration, we wanted to measure the impact of each of the features used by the system of~\newcite{shwartz2016improving}. Table \ref{tbl:results-ablation} presents the feature ablation results on the Alexa KB data using the fastText classifier. We compare the full set of features presented in section \ref{sct:isolating} against feature sets without the Brown clusters, word lemmas, POS tags, dependency information, and the $X$ and $Y$ entities (and their Brown cluster). We also show the results of the system using only the $X$ and $Y$ entities and just the words of the supporting sentences (without extracting the dependency path between entities).

The main takeaway is that for the \fact{instance of} and \fact{applies to} relations, the structure induced by the dependency parser is critical for the system's performance. One explanation is that these relations are not always lexically defined (sometimes expressed with just the verb `to be' across long subordinate clauses). For the \fact{birthplace of} relation, the system using the full sentence is on par with the best dependency-supported version suggesting that there are strong lexical cues that signify them (like `born in', or just the presence of a city name).

\section{Conclusion}
In this paper, we have explored the feature design and network architecture of the HypeNET RE system, and presented a new mechanism for extracting distant supervision data based on our large-scale KB. We found that by replacing HypeNET network architecture with a simple fastText model similar performance is achieved. The main difference between these two architectures is the mechanism of producing the high-dimensional representations: in HypeNET, LSTMs are used, which maintain dependency over longer contexts of dynamic length; in fastText, the window size for the ngrams is fixed. From our experiments, we can infer that dynamic-length context modelling did not bring any gains. Furthermore, we evaluated the effect of grouping of supports and satellites nodes features for various relations. The results from these experiments provide a solid ground to build RE systems for more relations. There are obvious extensions to the current approach, such as using a more sophisticated method for grouping the supports (e.g. an ensemble-based method) and we investigate these in future work.

Beyond architecture improvements, there are two main focus areas for the immediate future: generalising the system to cover very large number ($\sim$1k) of relations, and reducing the sources of noise. The former should be relatively straightforward given the existing architecture for extracting the dataset and training the system. The main obstacle will be to combine the results of the multiple RE systems (one for each relation group) into a single classifier.

Finally, distant supervision as a method itself introduces some errors since not all sentences that mention both entities of a fact express that fact (e.g. the relation \fact{is the director of} between \fact{steven spielberg} and \fact{saving private ryan} is not expressed in the sentence ``\textit{The level of violence in {Saving Private Ryan} makes sense because {Spielberg} is trying to show }\ldots''). %Perhaps the most effective way could be to obtain a manually annotated test set to correctly evaluate the performance of the system. 
Going further, we would like to expand the manual annotations to the training/validation sets to assist or replace the distant supervision.

\section{Bibliographical References}
\bibliography{references}
\bibliographystyle{lrec}
\end{document}